\documentclass[10pt,twocolumn,letterpaper]{article}

\usepackage{cvpr}              
\definecolor{cvprblue}{rgb}{0.21,0.49,0.74}
\usepackage[pagebackref,breaklinks,colorlinks,allcolors=cvprblue]{hyperref}
\usepackage{multirow}
\usepackage{multicol}
\usepackage{pifont}
\usepackage{colortbl} 
\usepackage{marvosym}
\usepackage[accsupp]{axessibility}

\def\paperID{} 
\def\confName{CVPR}
\def\confYear{2026}

\title{InstructTable: Improving Table Structure Recognition Through Instructions}

\author{
Boming Chen\\
Institution1\\
Institution1 address\\
{\tt\small firstauthor@i1.org}
\and
Second Author\\
Institution2\\
First line of institution2 address\\
{\tt\small secondauthor@i2.org}
}

\author{Boming Chen\textsuperscript{\rm 1}, Zining Wang\textsuperscript{\rm 1}, Zhentao Guo\textsuperscript{\rm 2}, Jianqiang Liu\textsuperscript{\rm 1}, Chen Duan\textsuperscript{\rm 1}, Yu Gu\textsuperscript{\rm 1},\\
    Kai Zhou\textsuperscript{\rm 1}\thanks{Corresponding Author}, Pengfei Yan\textsuperscript{\rm 1}\\
\textsuperscript{\rm 1} Meituan  \\
\textsuperscript{\rm 2} Beijing Institute of Technology\\
{\tt\small \{chenboming,wangzining03,zhoukai03\}@meituan.com, }
}

\begin{document}
\maketitle
\begin{abstract}
Table structure recognition (TSR) holds widespread practical importance by parsing tabular images into structured representations, yet encounters significant challenges when processing complex layouts involving merged or empty cells. Traditional visual-centric models rely exclusively on visual information while lacking crucial semantic support, thereby impeding accurate structural recognition in complex scenarios. Vision-language models leverage contextual semantics to enhance comprehension; however, these approaches underemphasize the modeling of visual structural information. To address these limitations, this paper introduces InstructTable, an instruction-guided multi-stage training TSR framework. Meticulously designed table instruction pre-training directs attention toward fine-grained structural patterns, enhancing comprehension of complex tables. Complementary TSR fine-tuning preserves robust visual information modeling, maintaining high-precision table parsing across diverse scenarios. Furthermore, we introduce Table Mix Expand (TME), an innovative template-free method for synthesizing large-scale authentic tabular data. Leveraging TME, we construct the Balanced Complex Dense Synthetic Tables (BCDSTab) benchmark, comprising 900 complex table images synthesized through our method to serve as a rigorous benchmark. Extensive experiments on multiple public datasets (FinTabNet, PubTabNet, MUSTARD) and BCDSTab demonstrate that InstructTable achieves state-of-the-art performance in TSR tasks. Ablation studies further confirm the positive impact of the proposed tabular-data-specific instructions and synthetic data. Code and datasets are available at \url{https://github.com/Nevar07/InstructTable}
\end{abstract}    
\vspace{-1.5em}
\section{Introduction}
\vspace{-0.5em}
\label{sec:intro}

Tables, as a fundamental medium for organizing and presenting structured information, are ubiquitous in academic publications, financial reports, official documents, and even natural scene images. Accurate \textbf{T}able \textbf{S}tructure \textbf{R}ecognition (TSR) which involves identifying cell boundaries, logical relationships (e.g., row/column merges), and spatial layouts, serves as a critical prerequisite to downstream tasks including data extraction, knowledge graph construction, and automated document analysis~\cite{icdar3}. However, due to the high heterogeneity of table layouts, such as empty cells, merged cells, and visual ambiguity, TSR remains a challenging task demanding methodological breakthroughs, particularly for complex tabular structures.

\begin{figure}[t]
\centering
\includegraphics[width=0.95\columnwidth]{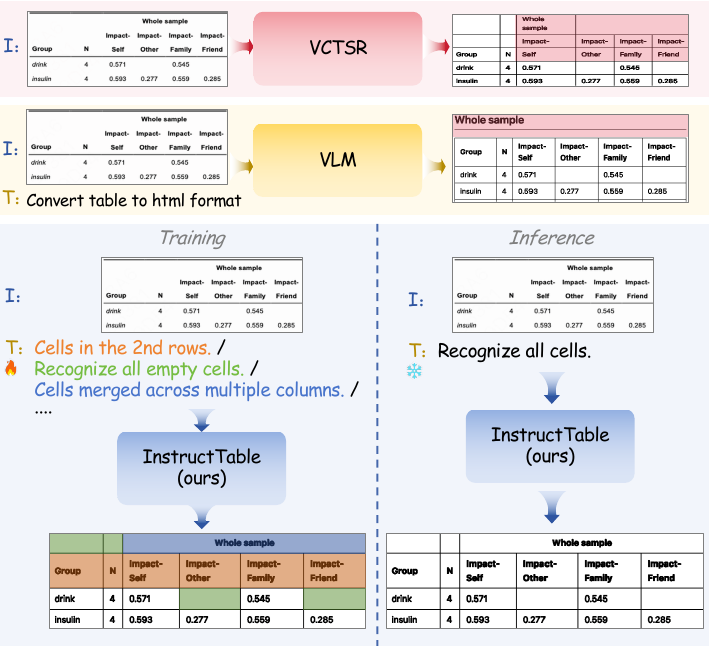} 
\caption{Visualized comparison among traditional \textbf{visual-centric TSR (VCTSR) models}, \textbf{vision-language models (VLM)}, and \textbf{InstructTable}. By leveraging instruction pre-training and TSR fine-tuning to jointly model visual information and instruction dependencies, InstructTable enhance fine-grained structural comprehension of tables.}
\label{fig:introduction}
\end{figure}

As illustrated in \cref{fig:introduction}, existing methods exhibit a semantic-visual imbalance: traditional visual-centric TSR models~\cite{pubtabnetedd,tablemaster,tableformer,mtltabnet,tablecenternet} solely rely on visual features while neglecting semantic associations, resulting in misjudgment of fine-grained table structure such merged cells or empty cells, whereas vision-language models~\cite{bai2025qwen25vl,seed16,comanici2025gemini,got,dots} overemphasize semantic information and compromise critical visual details such as alignment. To address this, we propose an instruction-guided TSR framework \textbf{InstructTable}, which dynamically modulates attention through multi-stage training strategy to collaboratively model visual structures and semantic dependencies.

\begin{figure}[t]
\centering
\includegraphics[width=0.95\columnwidth]{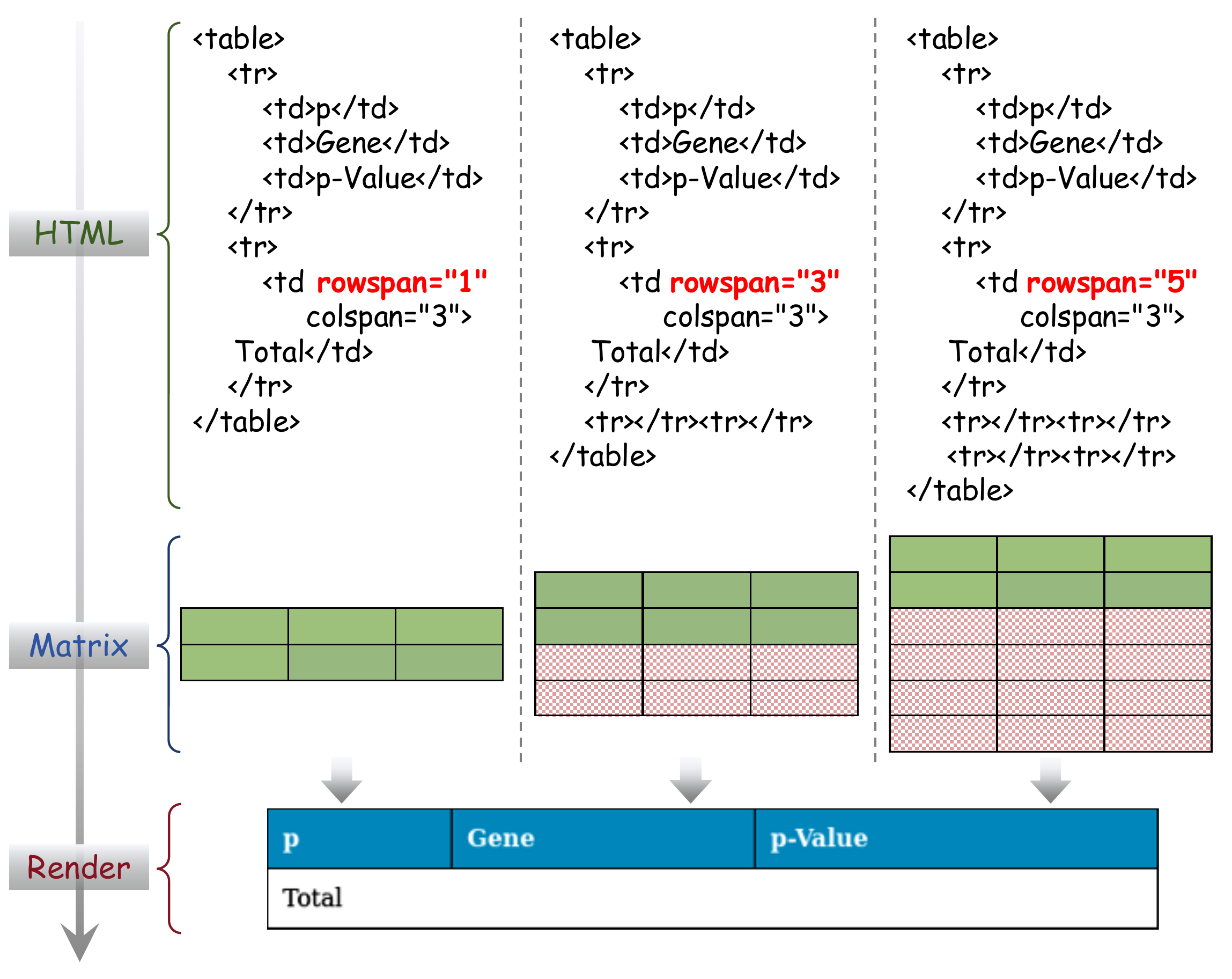} 
\caption{Implicit row problem visualization – identical images from divergent ground truths causing input-output misalignment, misleading model training. In the matrix, green regions denote normal cells while red dotted patterns indicate implicit rows.}
\label{fig:implicit_rows}
\end{figure}

Beyond model limitations, acquiring high-quality tabular data remains a significant challenge due to the complex annotation process including object detection, text recognition, and structural labeling. While existing method~\cite{tableformer} attempts to address this by synthesizing data through templates, the generated data exhibits limited diversity and authenticity. More critically, we identified systematic errors in existing template-based synthetic table datasets. Specifically, when cells with both \textit{rowspan} and \textit{colspan} attributes occupy full rows/columns, renderer generates implicit empty rows/columns during image generation. As illustrated in \cref{fig:implicit_rows}, this causes visual representations to contradict ground truth structural annotations, affecting about 2\% of the dataset and introducing over \textbf{20,000} redundant rows, compromising the dataset's utility for training. To resolve these limitations, we propose \textbf{T}able \textbf{M}ix \textbf{E}xpand (TME), a synthesis method that parses authentic tables into atomic cell matrices and subsequently splices them for data generation with the help of large language models (LLMs). While maximally preserving data authenticity, TME enables arbitrary mixing and expansion of tabular data. This approach effectively circumvents implicit structural errors and facilitates low-cost, large-scale generation of arbitrarily sized table datasets.

The primary contributions are summarized as follows:


\begin{itemize}
\item We introduce \textbf{InstructTable}, an instruction-guided multi-stage training TSR framework that bridges semantic instructions and visual information via instruction pre-training and TSR fine-tuning, significantly improving recognition accuracy for complex table structures.
\item We propose \textbf{TME}, parsing authentic tables into atomic cell matrix for arbitrary mixing and expansion. It enables large-scale, diverse synthetic tabular data generation from real samples, eliminating template-induced bias and alleviating high-quality data scarcity.
\item We introduce \textbf{B}alanced \textbf{C}omplex \textbf{D}ense \textbf{S}ynthetic \textbf{Tab}les (BCDSTab), a TME-synthesized benchmark with 900 complex table images for evaluating models in complex table scenarios.
\end{itemize}

\begin{figure*}[t]
\centering
\includegraphics[width=0.95\textwidth]{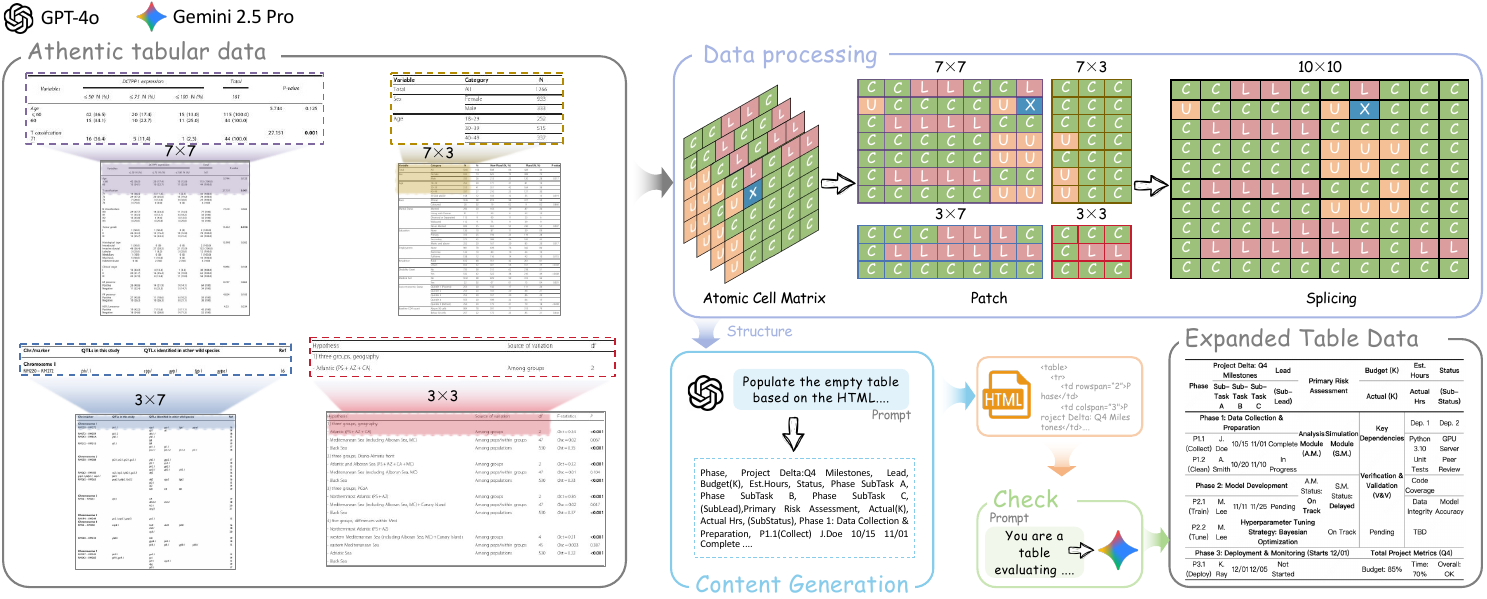} 
\caption{TME synthesis pipeline. Authentic table data undergoes matrix processing, partitioning, splicing, content generating and validity checking to generate novel synthetic tables for scenario-agnostic data expansion. In the atomic cell matrix, ``C" denotes independent cells, ``L" denotes left-merged cells, ``U" denotes up-merged cells, and ``X" denotes bidirectionally merged (left and up) cells.}
\label{fig:tme}
\end{figure*}

\section{Related Work}
\label{sec:related_work}

Advances in deep learning have driven the proliferation of TSR methods, which can be categorized into two paradigms: visual-centric and vision-language model based approaches, differentiated by their use of textual semantics.

\subsection{Visual-Centric Methods}
Visual-centric methods recognize table structures solely through visual information while disregarding semantic content, which typically results in their suboptimal performance in relatively complex scenarios. Based on the recognition pipeline, these methods can be broadly categorized into the following three classes.

\noindent \textbf{Split-and-Merge Methods.} These methods typically involve two phases: initial segmentation into row/column grids followed by cell merging. Early approaches like SPLERGE~\cite{SPLURGE} and SEM~\cite{sem} use semantic line segmentation with merge models. DeepDeSRT~\cite{deepdesrt} employs Faster R-CNN~\cite{fasterrcnn} and FCN~\cite{fcn} for joint detection/segmentation on electronic documents.
RobusTabNet~\cite{robusttabnet} introduces spatial CNN~\cite{scnn} for robust splitting. 
SEMv3~\cite{semv3} improves efficiency by predicting key point offsets on table lines. TABLET~\cite{tablet} formulates splitting as sequence labeling and merging as cell classification via transformer encoders. Due to their reliance on table lines, these methods demonstrate limited robustness in scenarios involving wireless tables, occlusions, creases, or similar challenging conditions.

\noindent \textbf{Detect-and-Classify Methods.} This group of methods operate by detecting individual cells and predicting their relationships to reconstruct table structures. Early approaches~\cite{clinchant2018comparing,xue2019res2tim,qasim2019rethinking} treat OCR text boxes as graph nodes, classifying relationships via graph neural networks. Subsequent methods focus on cell detection: Cycle-CenterNet~\cite{long2021parsing}, CascadeTabNet~\cite{prasad2020cascadetabnet}, and LGPMA~\cite{qiao2021lgpma} employ diverse techniques to enhance detection precision.
TGRNet~\cite{xue2021tgrnet} formulates relationship classification as logical link prediction using Graph Convolutional Networks~\cite{gcn}. 
LORE~\cite{lore} and LORE++~\cite{lore++} leverage cascaded transformers with multitask pretraining for physical/logical location prediction. 
These methods remain constrained by cell detection accuracy, often exhibiting degraded performance in wireless table and empty cell scenarios.

\noindent \textbf{Image-to-Markup Methods.} These methods directly generate structural sequences (e.g., HTML). EDD~\cite{pubtabnetedd} pioneers the encoder-dual-decoder paradigm: CNN feature extraction followed by RNN-based sequential generation of structure tokens and cell text. TableMaster~\cite{tablemaster} and TableFormer~\cite{tableformer} enhance decoding accuracy with Transformers, replacing text prediction with bounding box detection to reduce sequence length. VAST~\cite{vast} introduces Visual-HTML Alignment to align image regions with decoder features. 
UniTable~\cite{unitable} pretrains its encoder via VQ-VAE~\cite{vqvae} and generates sequences through three independent Transformers. SPRINT~\cite{sprint} deduces logical structures by predicting optimised table-structure language (OTSL) sequences~\cite{otsl} coupled with grid estimation~\cite{pubtables-1m}. Methods of this category are computationally intensive and data-hungry.

\subsection{Vision-Language Model Based Methods}
Vision-language models have demonstrated powerful capabilities across various image-text tasks~\cite{Guan_2025_ICCV,10887029,wang2025marten}. Beyond general-purpose large vision-language models  (e.g., GPT-4o~\cite{gpt4o}), several research efforts have explored task-specific fine-tuning for table scenarios. The OmniParser series~\cite{omniparser,omniparserv2} proposes a unified framework to simultaneously handle multiple text parsing tasks including TSR. TabPedia~\cite{tabpedia} integrates various visual table understanding tasks through a concept synergy mechanism. 
GOT~\cite{got} extends the OCR to recognize various optical signals, including tables, text, and other elements, under a unified character concept. 
MonkeyOCR~\cite{li2025monkeyocr} employs a Structure-Recognition-Relation (SRR) triplet paradigm, decomposing document parsing into layout analysis, content identification, and logical ordering to balance accuracy and efficiency. 
More recent work~\cite{niu2025mineru25,dots,cui2025paddleocr} employs a two-stage scheme of layout detection-content recognition for document parsing, while MinerU 2.5 and PaddleOCR-VL adopt OTSL for table representation to achieve higher efficiency and accuracy. 
Deepseek-OCR~\cite{wei2025deepseek} achieves high-accuracy table parsing with a reduced number of visual tokens through optical 2D mapping compression and an end-to-end VLM architecture.
Despite achieving superior performance in complex tabular scenarios through advanced semantic understanding, they may underperform visual-centric methods in scenarios requiring precise spacial information alignment.
\section{Methodology}

To address the dual challenges of scarce high-quality tabular data and models' imbalance between semantic and visual information, we propose: (1) TME, a template-free tabular data generation method based on real table structures and general LLM capabilities to generate tables at arbitrary scale; and (2) InstructTable, an end-to-end TSR approach leveraging instructions to enhance structural understanding.

\begin{figure*}[t]
\centering
\includegraphics[width=0.95\textwidth]{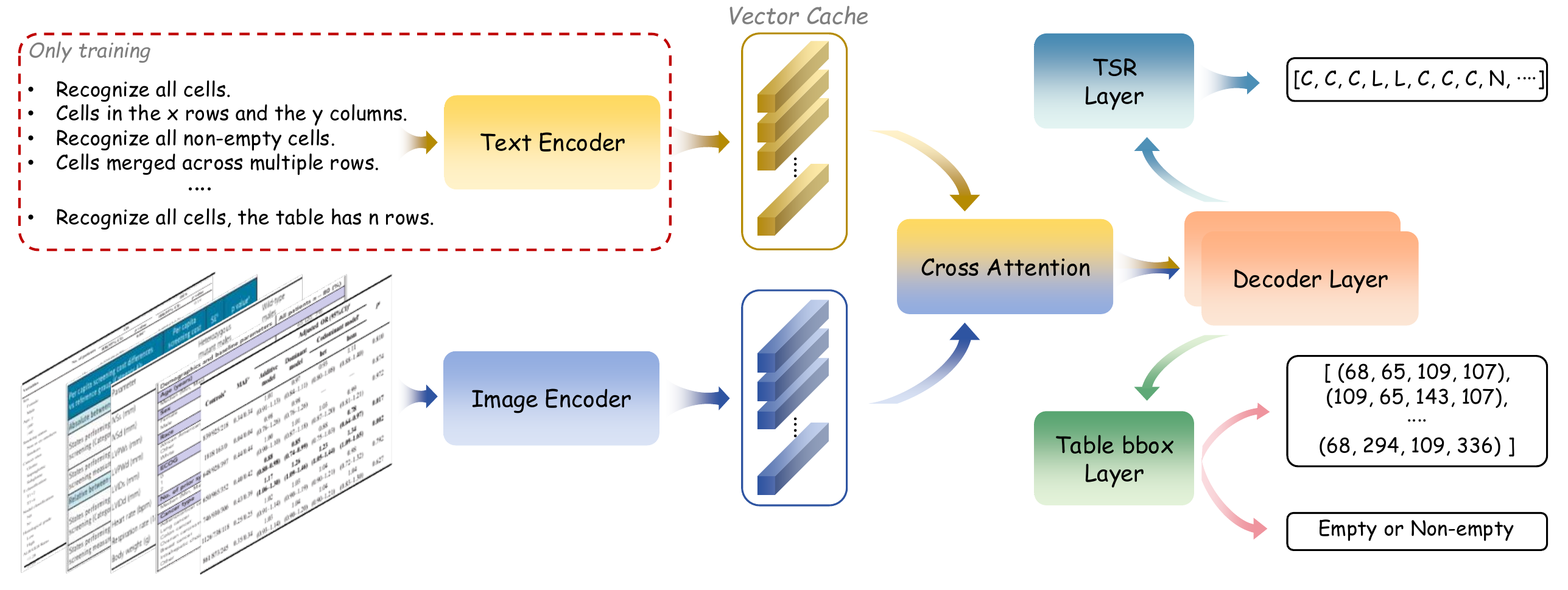} 
\caption{Main framework of InstructTable. The red dashed region indicates text encoding occurs only in training, where instruction embeddings are cached. During inference, cached vectors enable efficient processing without real-time text encoding.}
\label{fig:instructtable}
\end{figure*}

\subsection{Table Mix Expand}
Inspired by OTSL, we employ an atomic cell matrix approach to generate template-free authentic tabular data at arbitrary scales. Compared to HTML, this representation reduces token counts by over 80\% and generation lengths by approximately 50\% while preserving layout flexibility. Crucially, we observe that atomic cell matrices exhibit cropping invariance: any top-left submatrix remains valid, maintaining content topology and converting conflict-freely to HTML. Leveraging this property, our method: 
1) initializes an empty matrix with randomized dimensions and partitions it into N blocks; 
2) populates each block by selecting real data, cropping dimensionally compatible submatrices and introducing additional cell merges; 
3) converts the matrix into a structural-token-only HTML table sequence, populates the table content using LLMs such as GPT-4o~\cite{gpt4o}, and evaluates the validity and contextual semantic coherence of the generated table with Gemini 2.5 Pro~\cite{comanici2025gemini}, iterating this process until successful validation is achieved; 
4) applies random CSS augmentation followed by image rendering. 
As illustrated in \cref{fig:tme}, this process generates tables that intrinsically preserve tabular structure authenticity while ensuring semantic coherence of table contexts, operating without template constraints.

\subsection{Multi-Stage Training Strategy}
A multi-stage training strategy is utilized to enhance joint modeling of visual information and instructional dependencies, thereby strengthening tabular parsing capabilities across diverse scenarios. The strategy primarily comprises initialization, instruction pre-training, and TSR fine-tuning. In initialization, the input instruction is fixed as ``Recognize all cells" with several TSR training epochs, enabling the model to acquire fundamental TSR competence for improved subsequent training. Subsequently, we introduce human-designed fine-grained table understanding instructions to initiate a comprehensive instruction pre-training. During this stage, the model's attention is directed by instructions toward diverse structural granularities, significantly enhancing perception capabilities for complex tabular scenarios including merged cells and empty cells.Finally, we execute domain-adaptive fine-tuning on downstream benchmarks to augment comprehensive visual representation learning, which is aligned with established VCTSR paradigms.

\begin{table}[h]
\caption{Instruction templates for TSR in InstructTable. $n$ and $m$ denote table rows and columns respectively; $R$ and $C$ be subsets of $\{1, 2, \ldots, n\}$ and $\{1, 2, \ldots, m\}$, representing the randomly selected rows and columns; $x \in \{1, 2, \ldots, n\}$ and $y \in \{1, 2, \ldots, m\}$ indicate randomly chosen single row or column indices, ensuring high instruction diversity.}
\label{tab:instructions}
\centering
\resizebox{.45\textwidth}{!}{
\begin{tabular}{l|l}
\toprule
    \textbf{Group} & \textbf{Templates} \\
\hline
    \multirow{4}{0mm}{(1)} 
    & \textit{Recognize all cells.} \\
    & \textit{Recognize all cells, the table has $n$ rows.} \\
    & \textit{Recognize all cells, the table has $m$ columns.} \\
    & \textit{Recognize all cells, the table has $n$ rows and $m$ columns.} \\
\hline
    \multirow{4}{0mm}{(2)} 
    & \textit{Cells in the $R$ rows.}\\
    & \textit{Cells in the $C$ columns.} \\
    & \textit{Cells in the $x$ row and the $y$ column.} \\
    & \textit{Cells around the cell in the $x$ row and the $y$ column.} \\
\hline
    \multirow{2}{0mm}{(3)} 
    & \textit{Recognize all empty cells.} \\
    & \textit{Recognize all non-empty cells.} \\
\hline
    \multirow{3}{0mm}{(4)} 
    & \textit{Cells merged across multiple rows.} \\
    & \textit{Cells merged across multiple columns.} \\
    & \textit{Cells merged across multiple rows and multiple columns.} \\
\bottomrule
\end{tabular}}
\end{table}

\vspace{-1em}

\subsection{Instructions Design}
Building on prior findings that text-image fusion modules align linguistic instructions with visual information to enhance OCR performance~\cite{instructocr,docowl}, we design four groups of table-specific instructions: (1) recognition of table structure, (2) recognition of cells at specific locations, (3) recognition of empty cells, and (4) recognition of merged cells. The specific instructions for each group are comprehensively detailed in \cref{tab:instructions}. These linguistic instructions enable context-sensitive structural parsing while addressing tabular complexities, including structural nuances like empty and merged cells.

During instruction pre-training, instructions are randomly selected. Each training sample $x_i$ represented as a triplet $\{t_i, s_i, o_i\}$ where $t_i$ denotes the instruction text, $s_i$ represents the source image, and $o_i$ is the target output. For every source image $s_i$, we apply the control instruction $t_i$ to filter valid instances into the candidate set $O' = \{o'_0, o'_1, \ldots, o'_{n-1}\}$, then randomly sample a target instance $o_i \in O'$ as the generation objective conditioned on $(t_i, s_i)$. During prediction, the instruction is unified as: ``Recognize all cells''.

\begin{figure*}[t]
\centering
\includegraphics[width=0.95\textwidth]{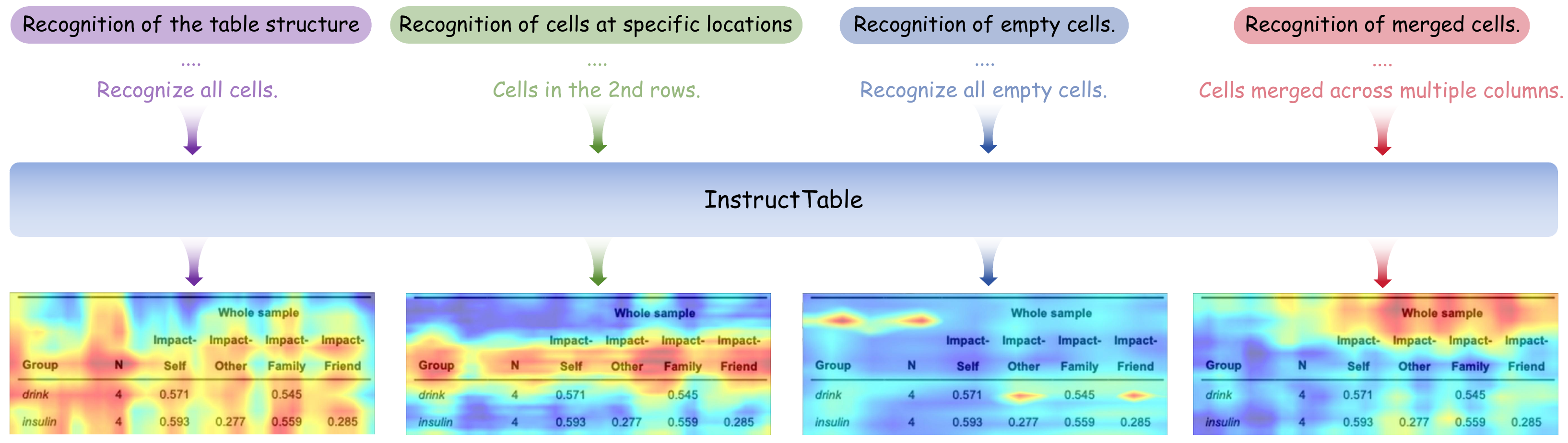} 
\caption{Attention heatmaps of the cross-attention layer under four instruction groups. Visualizations reveal how task-specific instructions dynamically modulate attention focus across table regions during parsing.}
\label{fig:attention}
\end{figure*}

\subsection{Model Structure}
We propose InstructTable, an end-to-end TSR framework that leverages instructions to modulate attention to enhance perception of fine-grained table structures. This approach improves structural understanding without introducing extraneous semantics. As illustrated in \cref{fig:instructtable}, the model jointly processes instructions and images to simultaneously generate three outputs: atomic cell matrices, cell bounding boxes, and empty cell classifications. 

\noindent \textbf{Image Encoder.}
The image encoder employs TableResNetExtra~\cite{tablemaster}, a table-optimized CNN backbone with 26 convolutional layers and 3 GCA blocks to enhance tabular feature perception. This architecture processes $960\times960$ input images into $60\times60$ feature maps, augmented with 2D positional embeddings before flattening into 1D vectors for cross-attention modules.

\noindent \textbf{Text Encoder.}
During training, the text encoder utilizes BERT~\cite{bert} and a liner projector to encode human-designed instructions into 512-dimensional feature vectors. These text embeddings are fused with image features through cross-attention to produce instruction-conditioned representations. For inference efficiency, we precompute and cache text embeddings after training, since the input instructions remain fixed during prediction.

\noindent \textbf{Decoder.}
The InstructTable decoder, built upon Transformer architecture, integrates three core components: instruction interaction via cross-attention, TSR layer generating atomic cell matrices, and bounding box prediction comprising coordinate regression and binary classification for empty cell detection. Offline OCR-derived text coordinates and content are fused with model output via matching rules to produce complete HTML representations, enabling efficient and accurate multilingual table parsing.




\subsection{Loss Function}
Multi-task learning is utilized on InstructTable to fit the model output to the corresponding data. For atomic cell matrix prediction and empty cell classification, a standard cross-entropy loss is used.

\begin{equation}
{\mathcal{L}_{seq}} = -\sum_{i=1}^{N} y_i \log(p_i) 
\label{equa: cross}
\end{equation}


\begin{equation}
{\mathcal{L}_{box\_cls}}=-\frac{1}{N}\sum_{i=1}^{N}\left [ {y}_{i}\log({p}_{i}) + (1-{y}_{i})\log(1-{p}_{i})\right ]  
\label{eq:bce}
\end{equation}

where $N$ denotes the number of samples in the batch, $y_i$ represents the true labels of the samples, and $p_i$ is the predicted probability distribution of the labels by the model. It is noteworthy that the empty cell classification is a binary classification task.

Moreover, our pipeline requires predicting the location of each cell to match with OCR results. Therefore, the L1 loss is used to constrain the position prediction, which is defined as follows:

\begin{equation}
\mathcal{L}_{box\_reg} = \sum_{i=1}^{N} \| x_i - \hat{x}_i \|_1
\label{equa: all}
\end{equation}

where $x_i$ denotes the ground truth position, $\hat{x}_i$ represents the predicted position.

Thus, the overall training loss is represented as follows:

\begin{equation}
\mathcal{L}_{total} = \alpha\mathcal{L}_{seq} +\beta\mathcal{L}_{box\_reg} + \gamma\mathcal{L}_{box\_cls}
\label{equa: all}
\end{equation}

where $\alpha$, $\beta$, and $\gamma$ are set to 1, 1, and 1, respectively.

\section{Experiments}

\begin{table*}[t]
\caption{TSR results on various datasets, respectively. InstructTable results are derived from pre-training on mixed data and fine-tuning on target datasets. Bolds denote the best and underlines denote the secondary best. $^{*}$ indicates results calculated by us from official checkpoints or online APIs. $\dagger$ denotes no incorporation of TME-generated data during the entire training process.}
\vspace{-0.5em}
\label{tab:results}
  \setlength\tabcolsep{2.6mm}
  \begin{tabular}{ccccccccc}
    \toprule
    \multirow{2}{*}{Methods} &  \multirow{2}{*}{Param} & \multicolumn{2}{c}{FinTabNet}  & \multicolumn{2}{c}{PubTabNet} & MUSTARD & \multicolumn{2}{c}{BCDSTab} \\
    \cmidrule(lr){3-4} \cmidrule(lr){5-6} \cmidrule(lr){7-7} \cmidrule(lr){8-9}
     & & S-TEDS & TEDS & S-TEDS & TEDS & S-TEDS & S-TEDS & TEDS \\
    \midrule
    \multicolumn{9}{c}{General Vision-Language Models} \\
    \midrule 
    GPT-4o$^{*}$~\cite{gpt4o} & - & 89.02 & 81.30 & 87.05 & 74.68 & 89.13 & 62.70 & 47.17  \\
    GPT-4.1$^{*}$~\cite{gpt41} & - & 90.75 & 84.64 & 87.61 & 77.14 & 91.37 & 74.88 & 56.75 \\
    Qwen2.5-VL-72B$^{*}$~\cite{bai2025qwen25vl} & 72B & 87.75 & 79.86 & 85.95 & 76.36 & 89.00 & 78.88 & 68.36  \\
    Qwen3-VL-235B-A22B$^{*}$~\cite{qwen3vl} & 235B & 92.26 & 88.36 & 91.75 & 85.37 & 92.94 & 81.41 & 72.31 \\
    Seed1.5-VL$^{*}$~\cite{guo2025seed} & 20B & 84.46 & 68.14 & 84.85 & 68.55 & 90.69 & 70.07 & 62.03  \\
    Seed1.6$^{*}$~\cite{seed16} & 230B & 85.31 & 71.74 & 86.14 & 70.13 & 93.35 & 78.97 & 62.55  \\
    InternVL3-78B$^{*}$~\cite{zhu2025internvl3} & 78B & 90.07 & 76.44 & 88.17 & 75.90 & 88.31 & 69.68 & 46.51 \\
    Claude 3.7 Sonnet$^{*}$~\cite{claude37} & - & 92.93 & 81.98 & 87.93 & 72.68 & 91.28 & 70.27 & 62.05  \\
    Gemini 2.5 Pro$^{*}$~\cite{comanici2025gemini} & - & 94.40 & 87.46 & 92.95 & 88.47 & 93.65 & 79.51 & 70.07 \\
    
    \midrule
    \multicolumn{9}{c}{Table-Aware Vision-Language Models} \\
    \midrule 
    OmniParser~\cite{omniparser} & - & 90.45 & 88.83 & 91.55 & 89.75 & - & - & -  \\
    TabPedia~\cite{tabpedia} & 7B & 95.41 & - & 95.11 & - & - & - & -  \\
    OmniParser V2~\cite{omniparserv2} & - & 93.20 & 90.50 & 90.50 & 88.90 & - & - & -  \\
    MonkeyOCR$^{*}$~\cite{li2025monkeyocr} & 3B & 89.56 & 81.40 & 94.61 & 90.08 & 69.86 & 76.16 & 67.90  \\
    dots.ocr~\cite{dots} & 1.7B & 87.86 & 84.12 & 93.76 & 90.65 & 90.92$^{*}$ & 78.43$^{*}$ & 71.28$^{*}$  \\
    MinerU2.5~\cite{niu2025mineru25} & 1.2B & 97.61 & 95.97 & 93.11 & 89.07 & 77.02$^{*}$ & 82.72$^{*}$ & \underline{74.07$^{*}$}  \\
    PaddleOCR-VL$^{*}$~\cite{cui2025paddleocr} & 0.9B & 95.03 & 93.12 & 89.82 & 85.09 & 67.65 & 78.08 & 72.71  \\
    DeepSeek-OCR$^{*}$~\cite{wei2025deepseek} & 3B & 96.48 & 95.43 & 90.69 & 86.44 & 80.93 & 62.98 & 48.72  \\
    
    \midrule
    \multicolumn{9}{c}{Visual-Centric TSR Models} \\
    \midrule
    EDD~\cite{pubtabnetedd} & - & 90.60 & - & 89.90 & 88.30 & - & - & - \\
    TableMaster~\cite{tablemaster} & 68M & 98.32 & 97.19 & 96.04 & 96.16 & 73.21$^{*}$ & 59.17$^{*}$ & 37.63$^{*}$ \\
    TableFormer~\cite{tableformer} & - & 96.80 & - & 96.75 & 93.60 & - & - & - \\
    TableFormer+OTSL~\cite{otsl} & - & - & 95.90 & - & 95.50 & - & - & - \\
    LORE~\cite{lore} & 23M & - & - & 98.10 & - & - & - & - \\
    VAST~\cite{vast} & - & 98.63 & 98.21  & 97.23 & 96.31 & - & - & - \\
    MTL-TabNet~\cite{mtltabnet} & 76M & 98.79 & - & 97.88 & 96.67 & 74.07 & - & - \\
    GridFormer~\cite{lyu2023gridformer} & - & 98.63 & - & 97.00 & 95.84 & - & - & - \\
    UniTable~\cite{unitable} & 382M & 98.89 & - & 97.89 & 96.50 & 84.14$^{*}$ & 76.22$^{*}$ & 60.94$^{*}$ \\
    SPRINT~\cite{sprint} & 80M & 98.03 & - & 95.71 & - & 85.19 & - & - \\
    TABLET~\cite{tablet} & - & 98.71 & 98.54 & 97.67 & 96.79 & - & - & - \\

    \midrule
    \rowcolor{blue!5}
    \textbf{InstructTable$\dagger$} & 69M & \underline{99.03} & \underline{98.57}  & \underline{98.11} & \underline{97.26} & \underline{94.21} & \underline{83.39} & 72.88 \\
    \rowcolor{blue!12}
    \textbf{InstructTable} & 69M & \textbf{99.20} & \textbf{98.68} & \textbf{98.22} & \textbf{97.46} & \textbf{95.30} & \textbf{84.75} & \textbf{75.45}  \\
    \bottomrule
  \end{tabular}
\end{table*}

\subsection{Datasets and Evaluation Metrics}

We train and evaluate InstructTable on three public benchmarks: PubTabNet~\cite{pubtabnetedd}, FinTabNet~\cite{fintabnet}, and MUSTARD~\cite{sprint}. These datasets collectively encompass diverse table layouts and multilingual content. PubTabNet provides 568K scientific publication tables with ground-truth HTML structural sequences, cell contents, and bounding box annotations. FinTabNet comprises 113K financial tables from S\&P 500 annual reports in PDF format, featuring comparable structural and cell-level annotations. MUSTARD serves as an evaluation-only benchmark comprising 1,428 multilingual tables spanning 13 languages. Sourced from scanned/printed documents and scene images, all tables feature OTSL-formatted logical structures.

In addition, we report the results on our proposed benchmark, BCDSTab. BCDSTab comprises 900 table images synthesized via TME based on PubTabNet and FinTabNet. BCDSTab provides comprehensive annotations: HTML structures, atomic cell matrix representations, nested cell/content bounding boxes, and text content. Noting that existing public datasets predominantly feature small-scale tables with limited cell counts, they inadequately represent prevalent real-world long-form tables. This sparsity creates academia-industry misalignment. BCDSTab bridges this gap by balancing data distributions and accounting for real-world size variations, establishing a challenging yet practical benchmark for TSR. Benchmark details are provided in Section 1 of supplementary.

We employ Tree-Edit-Distance-based Similarity (TEDS)~\cite{pubtabnetedd} and S-TEDS~\cite{vast} to evaluate end-to-end and structure-only table recognition performance. These metrics represent table structures as HTML trees, computing similarity scores via normalized tree-edit-distance between prediction and ground truth. In structure-only evaluation scenarios (e.g., MUSTARD), we exclusively utilize S-TEDS, which excludes textual content matching.

\subsection{Implement Details}
For each training dataset, 50k synthetic data is generated by TME for training augmentation. Each generated table receives randomly sampled row and column dimensions between 4 and 20, partitioned into 4 patches, and populated by authentic data from the corresponding dataset only.

The decoder comprises 4 transformer decoder layers with 8 attention heads. Output sequences have the maximum length of 1024, representing tables up to approximately $30\times30$ cells. The input image size is $960\times960$, with model dimension fixed at 512.

InstructTable is trained distributively across 8 NVIDIA A100-80G GPUs using Adam optimizer. Both initialization and instruction pre-training stages utilize a combined dataset of PubTabNet, FinTabNet, and TME-generated data, with global batch size 160. The initialization stage employs a fixed learning rate of 0.008 for 10 epochs. During instruction pre-training, an initial learning rate of 0.01 is applied with step decay scheduling across 200 total epochs. Following sufficient pretraining, 50 epochs of TSR fine-tuning are conducted on each downstream dataset, maintaining identical configurations as the initialization stage.

\subsection{Comparison with the State-of-the-Art Methods}

\noindent \textbf{Results of TSR.} 
To ensure a fair comparison, we conduct two separate experiments on InstructTable: one using only public datasets during the entire training process and another combining public datasets with TME-generated data. All results evaluated by us are obtained from official checkpoints or online APIs. Comprehensive experimental details are provided in Section 2 of supplementary.

As demonstrated in \cref{tab:results}, InstructTable achieves superior performance compared to all previous methods on FinTabNet, PubTabNet and MUSTARD. FinTabNet and PubTabNet containing predominantly English electronic documents, traditional visual-centric models outperform VLMs due to the relative simplicity of these scenarios and sufficient fine-tuning. When trained solely on public datasets, InstructTable shows comprehensive improvements over previous state-of-the-art models on both datasets. The incorporation of TME-generated data yields further performance gains, demonstrating that both proposed enhancements effectively strengthen the model's perception of table structures. On MUSTARD, characterized by its linguistic diversity and scenario richness, traditional visual-centric models struggle to achieve competitive performance. However, vision-language models, which are pre-trained on large-scale multimodal corpora, leverage their multilingual understanding capabilities to deliver superior results. Our evaluation shows that InstructTable achieves a TEDS score of 95.30\% on MUSTARD, surpassing the previous state-of-the-art visual-centric method SPRINT by 10.11\% and outperforming the strongest VLM Gemini 2.5 Pro by 1.65\%. This robust performance across varied scenarios demonstrates the effectiveness of our approach.


On our BCDSTab benchmark for complex long tables, we evaluated several open-source and API-based solutions. When trained solely on public datasets, InstructTable outperformed competitors in S-TEDS but trailed MinerU 2.5 in TEDS by 1.19\%. 
This gap stems from the domain discrepancy between simple public data and the complex scenarios in BCDSTab, which means that InstructTable never encountered lengthy table images during training. 
In contrast, MinerU2.5 undergoes extensive training on over 1M large-scal, high-quality table data. Coupled with its billion-scale parameters that confer powerful generalization capabilities, it achieves superior performance on the complex scenarios of BCDSTab.
After supplementing training with TME-generated data, InstructTable achieves a 2.57\% TEDS improvement on BCDSTab, surpassing all existing solutions to set a new state-of-the-art.

Detailed visual analyses of different methods on various benchmarks are presented in Section 3 of supplementary.

\noindent \textbf{TME Effectiveness Analysis.}
To verify the impact of TME-generated data on structural recognition, we conduct comparative experiments using both UniTable and InstructTable on FinTabNet and PubTabNet. For experimental consistency, all models were trained from scratch exclusively on their corresponding datasets, following the specifications of the original studies. Identical training procedures and data configurations are maintained across the four experimental groups. As detailed in \cref{tab:able_tme}, both models exhibit performance improvements when trained with TME-generated data. Across both datasets, the TEDS metric improves by an average of 1.2\%, while S-TEDS exhibits a mean increase of 0.7\%.

\begin{table}[h]
\caption{Comparative results of UniTable and InstructTable. ``TME" refers to the incorporation of TME-generated data.}
\label{tab:able_tme}
\setlength{\tabcolsep}{3.3pt}
  \begin{tabular}{cccccc}
    \toprule
    \multirow{2}{*}{Method} & \multirow{2}{*}{TME} & \multicolumn{2}{c}{FinTabNet} & \multicolumn{2}{c}{PubTabNet} \\
    \cmidrule(lr){3-4} \cmidrule(lr){5-6}
    & & S-TEDS & TEDS & S-TEDS & TEDS \\
    \midrule
     Unitable & \ding{56} & 96.91 & 95.10 & 95.00 & 93.23 \\
     Unitable & \ding{51} & 97.75 & 95.85 & 96.09 & 94.83 \\
     InstructTable & \ding{56} & 98.66 & 96.82 & 96.84 & 94.97 \\
     InstructTable & \ding{51} & \textbf{98.78} & \textbf{97.65} & \textbf{97.51} & \textbf{96.63}  \\
    \bottomrule
  \end{tabular}
  
\end{table}

\subsection{Ablation Studies}

\noindent \textbf{Impact of Framework Components.}
In this section, we conduct a comparative analysis to evaluate the impact of the two core framework components: instruction pre-training and TME-generated data utilization. As evidenced in \cref{tab:able_model}, both components positively impact model performance on both datasets. Instruction pre-training significantly improves structural parsing accuracy, while TME data enhances cell localization precision, yielding greater TEDS improvements. When combined, these components demonstrate synergistic effects, achieving average gains of +1.6\% TEDS and +1.8\% S-TEDS across the two datasets.

\begin{table}[h]
\caption{Ablation study of our proposed components on FinTabNet and PubTabNet, ``INS" and ``TME" refer to utilization of instruction pre-training and the incorporation of TME-generated data.}
\label{tab:able_model}
\setlength{\tabcolsep}{6pt}
  \begin{tabular}{cccccc}
    \toprule
    \multirow{2}{*}{INS} & \multirow{2}{*}{TME} & \multicolumn{2}{c}{FinTabNet} & \multicolumn{2}{c}{PubTabNet} \\
    \cmidrule(lr){3-4} \cmidrule(lr){5-6}
    & & S-TEDS & TEDS & S-TEDS & TEDS \\
    \midrule
     \ding{56} & \ding{56} & 97.11 & 96.57 & 95.91 & 94.03 \\
     \ding{51} & \ding{56} & 98.66 & 96.82 & 96.84 & 94.97 \\
     \ding{56} & \ding{51} & 98.37 & 97.07 & 96.22 & 95.01 \\
     \ding{51} & \ding{51} & \textbf{98.78} & \textbf{97.65} & \textbf{97.51} & \textbf{96.63} \\
      & & (\textbf{+1.67}) & (\textbf{+1.08}) & (\textbf{+1.60}) & (\textbf{+2.60}) \\
    \bottomrule
  \end{tabular}
  
\end{table}

\noindent \textbf{Impact of Distinct Instruction Groups.}
A set of experiments are conducted to investigate the impact of distinct instruction groups on InstructTable. Each group is individually removed from the instruction library for experimentation. The final results are presented in \cref{tab:able_ins}. The removal of any single group leads to a slight performance decline, with the first group, which guides the model to recognize the overall table structure, exhibiting a more significant impact (2.41\% S-TEDS decrease). The results demonstrate that incorporating instructions for recognizing overall table structures is essential as they define the model's core output objective. Moreover, a sufficiently large and diverse instruction library facilitates balanced attention distribution and strengthen its understanding of fine-grained table structures.

\begin{table}[h]
\caption{Ablation study of instruction groups on PubTbaNet without further fine-tuning.
    ``G$_{1}$",``G$_{2}$",``G$_{3}$" and ``G$_{4}$" refer to the corresponding instruction groups.
  }
\label{tab:able_ins}
\setlength{\tabcolsep}{14pt}
  \begin{tabular}{ccccc}
    \toprule
    \multicolumn{4}{c}{Instruction Group} & PubTabNet \\
    \cmidrule(lr){1-4} \cmidrule(lr){5-5}
    G$_{1}$ & G$_{2}$ & G$_{3}$ & G$_{4}$ & S-TEDS \\
    \midrule
    \ding{56} & \ding{51} & \ding{51} & \ding{51} & 93.93 \\
    \ding{51} & \ding{56} & \ding{51} & \ding{51} & 96.21 \\
    \ding{51} & \ding{51} & \ding{56} & \ding{51} & 96.04 \\
    \ding{51} & \ding{51} & \ding{51} & \ding{56} & 96.05 \\
    \ding{51} & \ding{51} & \ding{51} & \ding{51} & \textbf{96.34} \\
    \bottomrule
  \end{tabular}
  
\end{table}

Furthermore, we present quantitative results demonstrating the impact of varying instructions. As visualized in \cref{fig:attention}, which depicts cross-attention heatmaps for the same image under different instructions, InstructTable dynamically allocates attention to instruction-relevant regions. Unlike text-spotting models that exhibit localized attention patterns, our approach requires global image context to comprehend table structures, even when locating partial cell subsets. Consequently, while concentrating on instruction-specified cells, the model maintains holistic image attention rather than restricting focus exclusively to target cells.

\section{Conclusion}
In this paper, we propose InstructTable, a novel instruction-guided multi-stage training TSR framework that leverages human-designed instructions to enhance the understanding of fine-grained table structures. Through three training stages, the model integrates our meticulously crafted tabular-data-specific instruction sets with visual information to mitigate the visual-semantic imbalance observed in previous TSR methods. In addition, we introduce TME, a template-free tabular data synthesis method. By parsing tables into atomic cell matrices followed by mixed expansion, then employing LLMs for content population and validation, it enables large-scale, automatic construction of high-quality tabular data while preserving authenticity. Furthermore, leveraging this approach, we propose the BCDSTab benchmark, comprising 900 balanced complex dense tables for TSR capability evaluation. Extensive experiments demonstrate state-of-the-art performance on multiple benchmarks, with particularly gains in complex scenarios.

\newpage
{
    \small
    \bibliographystyle{ieeenat_fullname}
    \bibliography{main}
}
\clearpage

\begin{figure*}[h]
\centering
\includegraphics[width=1\textwidth]{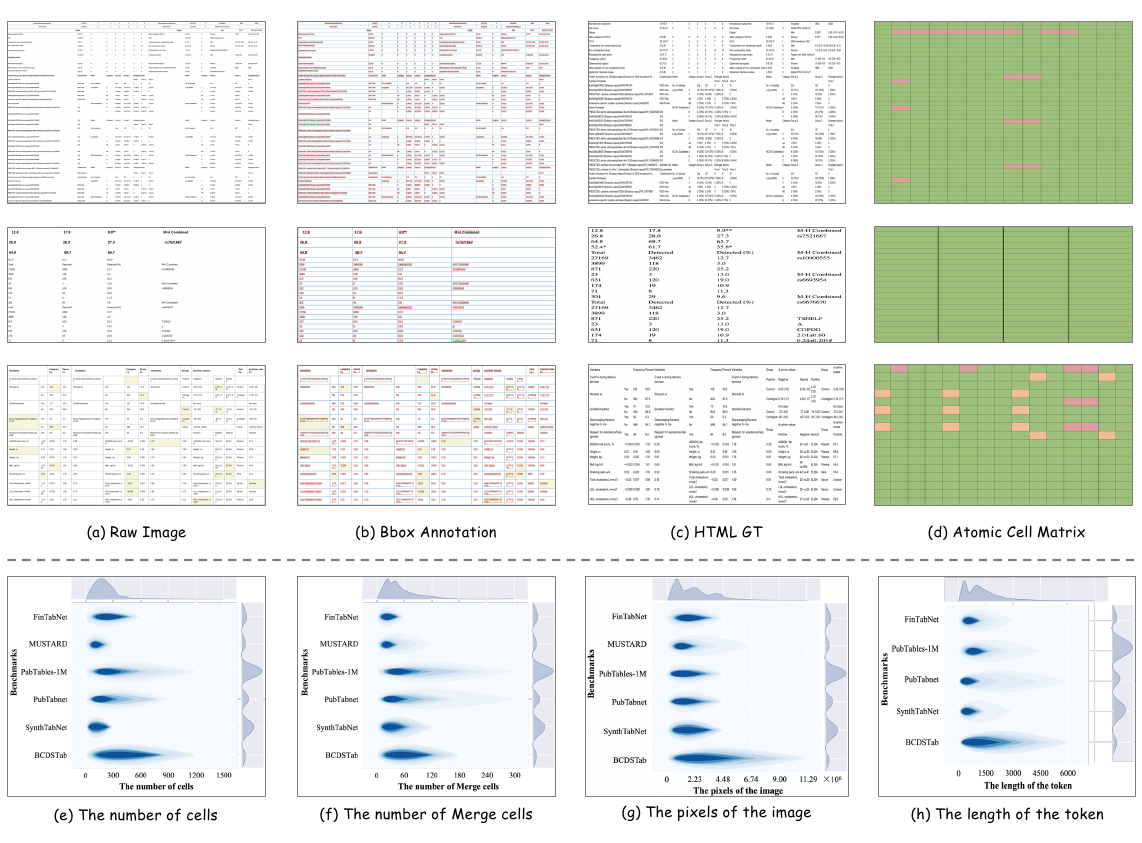}
\caption{An overview of BCDSTab benchmark. (a) Representative image sample; (b) Dual-level bounding box visualization with blue indicating cell-level boundaries and red showing content-level text regions; (c) HTML ground truth representation for metric evaluation; (d) Atomic cell matrix color-coded by merge patterns: green for normal cells, red for left-merged cells, yellow for up-merged cells; (e)-(h) Multidimensional data statistics.}
\label{fig:dataset}
\end{figure*}

\section{BCDSTab Benchmark Detailed Specifications}
The Balanced Complex Dense Synthetic Tables (BCDSTab) benchmark comprises 900 dense table images synthesized through our TableMixExpand (TME) framework, with source data derived from FinTabNet \cite{fintabnet} and PubTabNet \cite{pubtabnetedd}. During synthesis, we first sample the total cell count $C$ from a normal distribution $\mathcal{N}$ bounded within $[4, 1000]$. Next, we uniformly select the row count $R$ from discrete integers in $[2, 100]$. The column count $K$ is then computed via integer division: $K = \lfloor C / R \rfloor$. If $K$ falls outside the valid range $[2, 15]$, we iteratively resample $C$ and $R$ until satisfying $2 \leq K \leq 15$. This rejection sampling procedure initializes an empty atomic cell matrix ready for content generation. 

Subsequently, the initialized matrix is partitioned into four patches according to its row-column dimensions. We randomly sample four authentic tables from the hybrid corpus combining FinTabNet and PubTabNet. For each source table, a top-left submatrix matching the corresponding patch dimensions is extracted. When source dimensions are insufficient, tables are resampled until meeting size requirements. These submatrices are then populated into their respective patches within the target matrix, resulting in a populated atomic cell matrix with complete structural, while without any text content.

Upon obtaining the complete atomic cell matrix, it is converted into a minimal HTML representation containing only table structure elements. This processed HTML is then combined with the prompt ``Populate the empty table based on the HTML provided. Return a complete table! Ensure the table structure exactly match the empty table provided!'' and fed into GPT-4o~\cite{gpt4o}, leveraging the extensive knowledge of this general-purpose large language model to generate contextually appropriate textual content based on the table structure.Then we employ the powerful reasoning model Gemini-2.5 Pro~\cite{comanici2025gemini} to verify the synthetic table' validity and contextual coherence, using the prompt: ``You are a table evaluating expert, you will receive an HTML-formatted table to verify both its structural compliance and the contextual coherence of its content.''

We then apply randomized CSS augmentations to the HTML, incorporating stochastic variations in:
\begin{itemize}
    \item Text alignment direction (left/center/right/justify)
    \item Font families (12 serif/sans-serif options)
    \item Font sizes (range: 10-25pt)
    \item Vertical padding (2-12px)
    \item Border styles (solid/dashed/dotted/double/none)
    \item Table line types (single/double/hidden)
    \item Text colors
    \item Cell background colors
\end{itemize}

The augmented HTML is rendered on a Linux server using headless Chrome (v91.0.4472.77), constrained by maximum dimensions of 5000 pixels in height and 3000 pixels in width, with a minimum font height requirement of 12 pixels. Samples violating these constraints are discarded. Chrome's rendering engine enables precise element localization via XPath queries, yielding accurate pixel coordinates for all cell boundaries.

For non-empty cells, we perform text bounding box extraction through computer vision processing:
\begin{enumerate}
    \item Convert cell sub-image to binary format
    \item Detect text contours using OpenCV's \texttt{boundingRect}
    \item Map local coordinates to global image
    \item Output cell content bounding boxes in $[x_{min}, y_{min}, x_{max}, y_{max}]$ format matching FinTabNet/PubTabNet specifications
\end{enumerate}

The controlled color differentiation ensures reliable binarization. This process yields the complete BCDSTab benchmark, comprising 1,000 dense table images, HTML ground-truth representations, atomic cell matrix structures, cell-level bounding boxes, and content-level bounding boxes. Sample visualizations of the benchmark are presented in Figure \cref{fig:dataset}.

Furthermore, we conduct quantitative comparisons across multiple dimensions between BCDSTab and leading public table datasets. Specifically, FinTabNet, PubTabNet, PubTables-1M \cite{pubtables-1m}, SynThTabNet \cite{tableformer}, and MUSTARD \cite{sprint} with their test sets. As quantified in Figure \cref{fig:dataset}, BCDSTab demonstrates superior distribution balance and broader value ranges in both cell counts and merged cell quantities, fulfilling our objective to evaluate TSR models under dense, long-table scenarios. BCDSTab provides images with significantly higher pixel counts than existing datasets, delivering enhanced visual detail to support flexible downstream processing by users. 

To prevent truncation of vision-language model outputs, we analyze token counts of full HTML representations across datasets using the BERT tokenizer \cite{bert}. The maximum observed token length was 5867, prompting our configuration of the vision-language model's output limit at 8192 tokens, ensuring complete table generation capacity. Notably, MUSTARD is excluded from this analysis as it provides only structural annotations without text content.

\section{Vision-Language Model Experimental Setup}
In our experimental evaluation, we conduct extensive testing across multiple vision-language models while maintaining fixed configurations for each model to ensure fairness, with detailed settings specified as follows:

\begin{enumerate}
    \item We fix the prompt to ``This is an image containing only one table, please convert the table in the image to HTML (begin with $\langle \rm{table} \rangle$ and end with $\langle \rm{/table} \rangle$) format. Only the content in the image needs to be output without expanding other content.'' uniformly across all general VLMs.
    \item We set the maximum output length to 8192 tokens, which is sufficient for generating complete HTML table outputs across all datasets.
    \item We systematically employ regular expressions to extract clean table content enclosed within ``$\langle \rm{table} \rangle$'' and ``$\langle \rm{/table} \rangle$'' tags from model outputs, ensuring effective isolation from extraneous textual elements.
    \item For reasoning-capable models (e.g., Gemini 2.5 Pro), we consistently enable their think mode during inference. 
    \item For the multi-stage Table-aware VLMs (e.g., MinerU2.5~\cite{niu2025mineru25}), we isolate their table structure recognition module separately and employ official parameters and prompts to perform table structure recognition.
\end{enumerate}

\section{Case Study}

\begin{figure*}[t]
\centering
\includegraphics[width=1\textwidth]{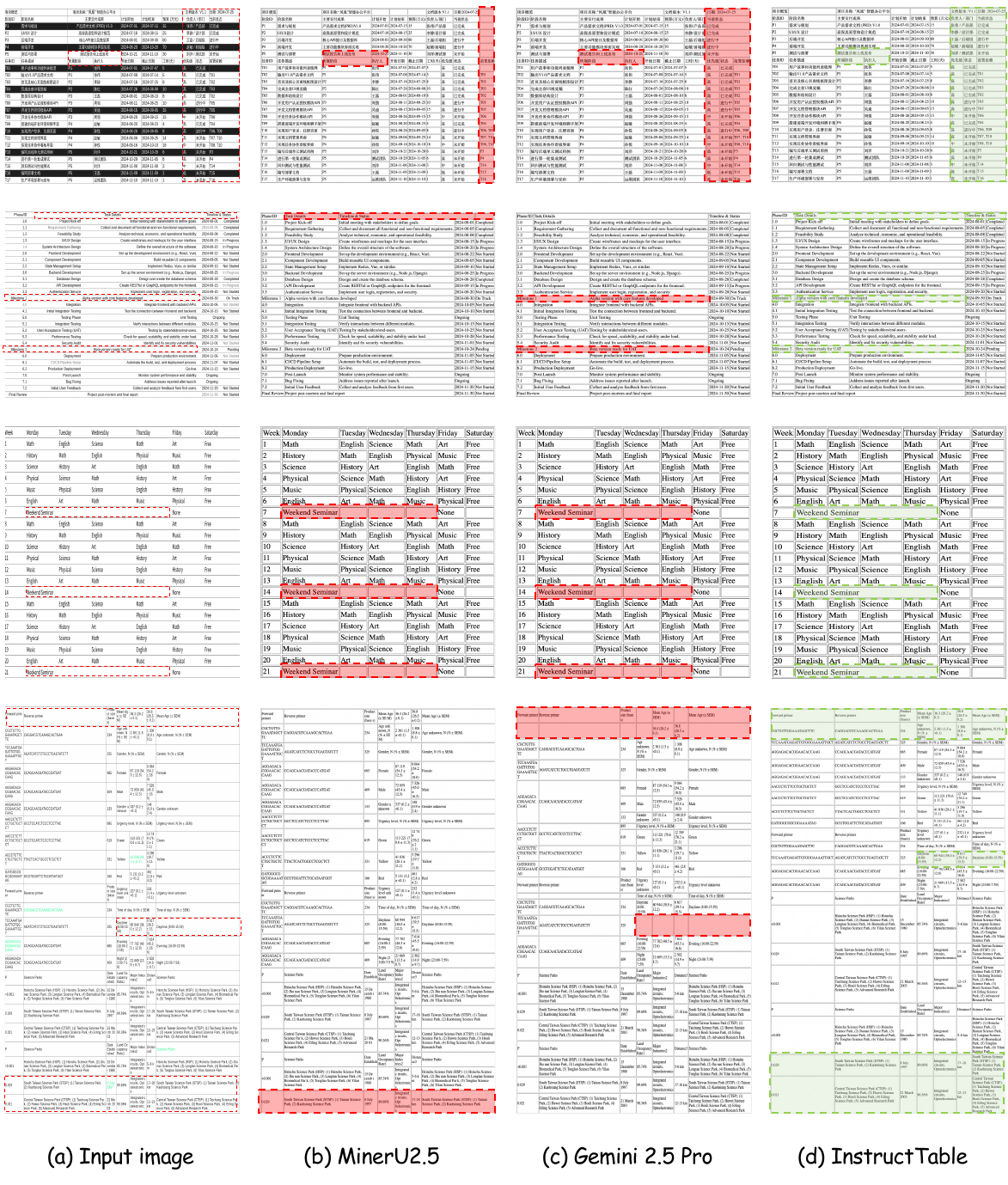}
\caption{Visual comparison of different methods on BCDSTab, including: (a) input image, (b) MinerU2.5 results, (c) Gemini 2.5 Pro results, and (d) our InstructTable results. Red regions highlight erroneous positions in the predictions, with corresponding areas marked by green zones in our predictions and delineated by red dashed boxes in the input image for intuitive comparison.}
\end{figure*}

\begin{figure*}[t]
\centering
\includegraphics[width=1\textwidth]{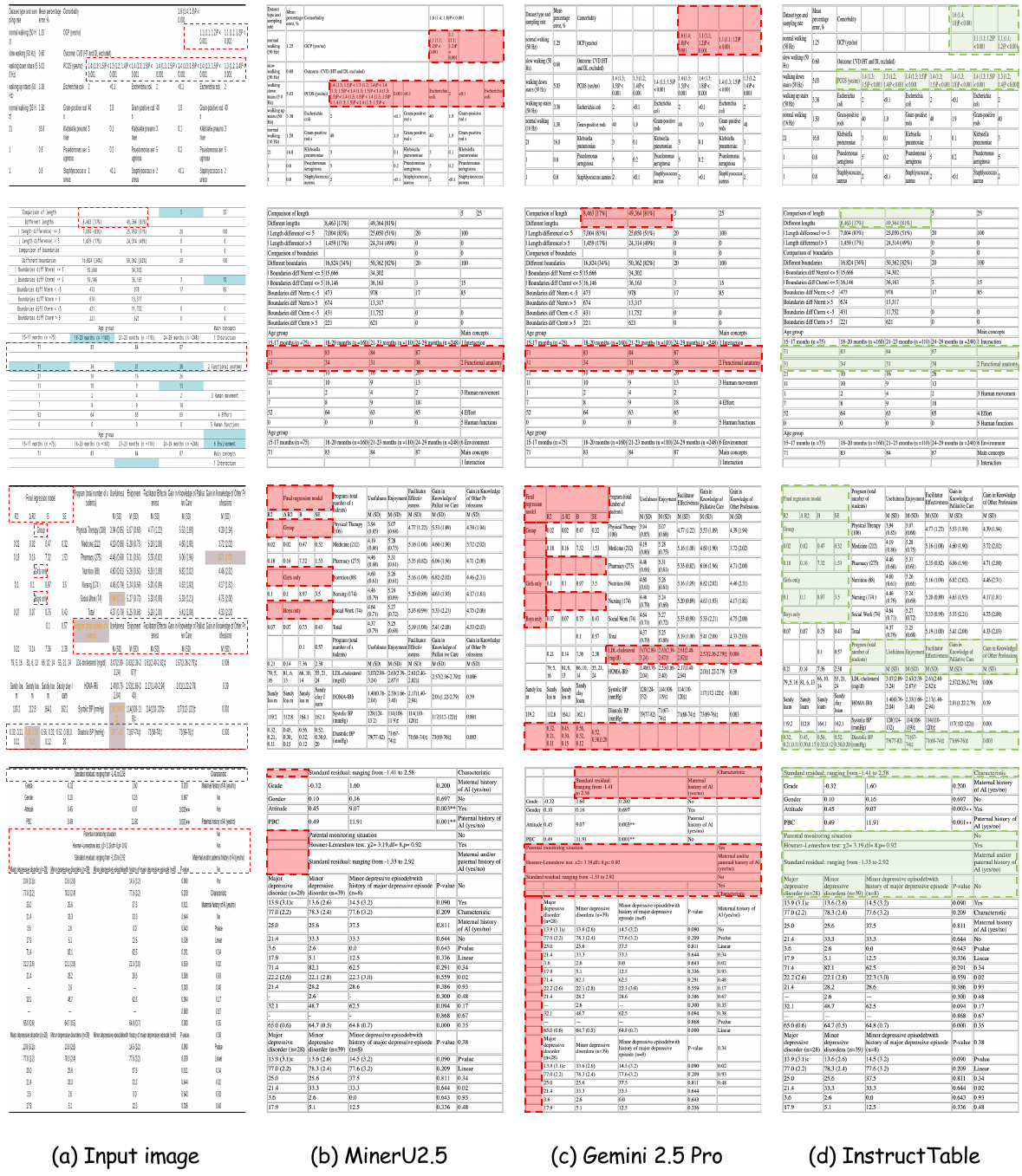}
\caption{Visual comparison of different methods on BCDSTab, including: (a) input image, (b) MinerU2.5 results, (c) Gemini 2.5 Pro results, and (d) our InstructTable results. Red regions highlight erroneous positions in the predictions, with corresponding areas marked by green zones in our predictions and delineated by red dashed boxes in the input image for intuitive comparison.}
\end{figure*}


\end{document}